\documentclass[tinyml]{acmart}
\usepackage[capitalise,nameinlink]{cleveref}
\usepackage{adjustbox}
\usepackage{subcaption}
\usepackage{xfrac}
\usepackage{xcolor}
\usepackage{tikz}
\usetikzlibrary{arrows,calc,positioning,backgrounds}

\definecolor{cgreen}{rgb}{0.0, 0.3, 0.0}

\AtBeginDocument{%
  \providecommand\BibTeX{{%
    \normalfont B\kern-0.5em{\scshape i\kern-0.25em b}\kern-0.8em\TeX}}}

\setcopyright{rightsretained}
\copyrightyear{2024}
\acmYear{2024}

\begin{document}

\title[Towards a tailored mixed-precision sub-8-bit quantization scheme for GRUs with GA]{Towards a tailored mixed-precision sub-8-bit quantization scheme for Gated Recurrent Units using Genetic Algorithms}

\author{Riccardo Miccini}
\orcid{0000-0002-0421-6170}
\email{rimi@dtu.dk}
\author{Alessandro Cerioli}
\email{alceri@dtu.dk}
\affiliation{%
  \institution{Technical University of Denmark}
  \city{Kongens Lyngby}
  \country{Denmark}}

\author{Clément Laroche}
\email{claroche@jabra.com}
\author{Tobias Piechowiak}
\email{topiechowiak@jabra.com}
\affiliation{%
 \institution{GN Audio}
 \city{Ballerup}
 \country{Denmark}}

\author{Jens Sparsø}
\email{jspa@dtu.dk}
\author{Luca Pezzarossa}
\email{lpez@dtu.dk}
\affiliation{%
  \institution{Technical University of Denmark}
  \city{Kongens Lyngby}
  \country{Denmark}}

% \author{Anonymous}
% \affiliation{%
%   \institution{Anonymous}}

\renewcommand{\shortauthors}{Miccini, et al.}
% \renewcommand{\shortauthors}{Anonymous, et al.}

\begin{abstract}
  Despite the recent advances in model compression techniques for deep neural networks, deploying such models on ultra-low-power embedded devices still proves challenging. 
  In particular, quantization schemes for Gated Recurrent Units (GRU) are difficult to tune due to their dependence on an internal state, preventing them from fully benefiting from sub-8bit quantization. 
  In this work, we propose a modular integer quantization scheme for GRUs where the bit width of each operator can be selected independently. 
  We then employ Genetic Algorithms (GA) to explore the vast search space of possible bit widths, simultaneously optimizing for model size and accuracy. 
  We evaluate our methods on four different sequential tasks and demonstrate that mixed-precision solutions exceed homogeneous-precision ones in terms of Pareto efficiency. 
  Our results show a model size reduction between $25\%$ and $55\%$ while maintaining an accuracy comparable with the 8-bit homogeneous equivalent.
  
\end{abstract}

\keywords{Neural networks, Quantization, Neural architecture search}

\maketitle

\section{Introduction}\label{sec:intro}

In recent years, deep neural networks have been making strides in a wide range of fields, including computer vision, natural language processing, and audio processing.
In particular, Recurrent Neural Networks (RNNs) --- specifically their Long Short-Term Memory (LSTM)~\cite{hochreiter_lstm_1997} and Gated Recurrent Unit (GRU)~\cite{cho_learning_2014} variants --- have been successfully applied to a variety of sequence modelling tasks, such as speech recognition, machine translation, keyword spotting, and speech enhancement. 
However, deploying such models on ultra-low-power embedded devices still proves challenging, due to their high computational complexity and memory footprint.
Specifically, data transfer --- and by extension model size --- is a major contributor to the overall energy footprint of a deep learning model~\cite{Horowitz_computing_2014}.

Model compression techniques such as quantization have been successfully applied to Convolutional Neural Networks (CNNs), allowing them to be deployed on embedded devices with limited computational resources.
Remarkably, the quantization of RNNs has not been explored as extensively, potentially due to the additional complexity introduced by their recurrent nature.
Among the most notable works, \cite{alom_effective_2018} propose binary, ternary, and quaternary quantization schemes for RNNs and evaluate it on sentiment analysis, \cite{fedorov_tinylstms_2020} combines structural pruning and 8-bit quantization to optimize LSTMs for speech enhancement on a Cortex-M7 embedded platform, \cite{li_quantization_2021} presents quantization schemes for the standard LSTM and its variants, based on fixed-point arithmetic, evaluating them on speech recognition; finally \cite{rusci_accelerating_2022} employs mixed-precision FP16 and 8-bit integer quantization to deploy speech enhancement models based on LSTMs or GRUs on a RISC-V embedded target. 

Neural Architecture Search (NAS) is a recently introduced technique for automating the design of neural networks.
It consists of exploring the search space of possible neural network architectures to optimize them according to one or several metrics, such as model accuracy, size, or computational complexity.
Within NAS, several methods for traversing the vast search space of possible architectures have been proposed, including gradient-based methods~\cite{santra2021gradient}, Reinforcement Learning~\cite{chitty2022neural}, and Genetic Algorithms (GA). 
In particular, the latter has been adopted in computer vision as a way to find an optimal CNN structure for face recognition~\cite{rikhtegar2016face} and, more recently, achieving better performances than manually-derived CNNs for image classification~\cite{real2019regularized}. 
Furthermore, GA have been demonstrated to be effective in the refinement of RNNs for natural language processing~\cite{klyuchnikov2022bench}. 
Vector extensions of genetic algorithms have also been proposed: \cite{lu_nsganet_2019} proposes NSGA-Net, a multi-objective GA for optimizing neural networks for image classification, while \cite{termritthikun2021neural} uses GA to simultaneously optimize inference time and model size for anomaly detection. 
For a comprehensive overview of the field, we refer the reader to \cite{benmeziane_comprehensive_2021,kang2023neural}.

Nevertheless, most NAS solutions focus on optimizing architectural aspects of the network such as the number of layers or the type of activation functions, whereas the hyperparameters associated with quantization are often handled manually.
This can be particularly strenuous, especially in mixed-precision quantization settings where the search space of quantization parameters is vast.
Addressing this gap, we contribute to the field of quantization by employing GA to efficiently derive mixed-precision quantization schemes for GRUs which simultaneously maximizes accuracy and minimizes model size.
Such a quantization scheme is particularly relevant for embedded devices with limited storage, memory, or computational resources, where model weights and activation size are critical factors.
In this work we contribute by:
\begin{itemize}
  \item Designing an integer-only modular quantization scheme for GRUs that can be used for sub-8-bit computation;
  \item Employing Genetic Algorithms to derive Pareto-optimal mixed-precision quantization schemes;
  \item Conducting a comprehensive evaluation of the solution, showcasing its effectiveness across four diverse sequence classification tasks of varying complexity.
\end{itemize}

\section{Background}\label{sec:bg}
This section provides a theoretical background on the main building blocks of this work, namely GRUs, quantization, and GA.

\subsection{Gated Recurrent Unit}\label{ssec:gru}
Recurrent Neural Networks, are a class of neural networks first introduced in \cite{rumelhart_learning_1986} as a way to process sequential data of arbitrary length.
RNNs process data sequentially over several time steps: at each time step, the network takes as input the current data and the previous output, allowing the network to maintain an internal state.
Because of this, RNNs are well suited for processing data such as text or audio.
However, in their most basic form, RNNs suffer from the so-called \textit{vanishing gradient problem}~\cite{bengio_learning_1994}, which makes them difficult to train on long sequences.

To address this issue, LSTM units~\cite{hochreiter_lstm_1997}, and subsequently GRUs~\cite{cho_learning_2014}, were introduced.
Both LSTMs and GRUs implement a number of gates to modulate the flow of new and past information, allowing a more compact internal representation that does not suffer from vanishing gradient.

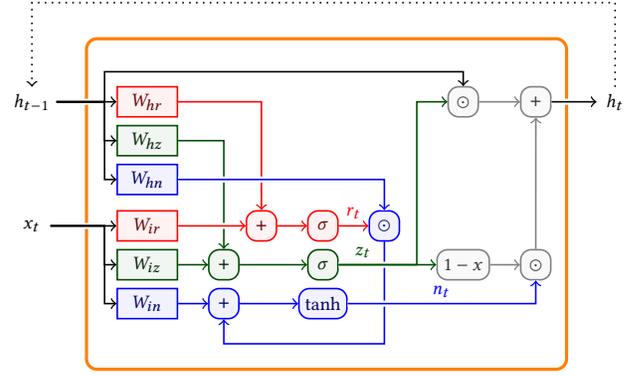
\begin{figure}[t]
  \centering
  \begin{tikzpicture}[node distance=1mm and 4mm, auto, scale=0.8, every node/.style={scale=0.8}]
    % Styles
    \tikzstyle{inout} = [minimum width=6mm]
    \tikzstyle{fc}  = [rectangle, draw, semithick, text centered, minimum height=5mm, minimum width=1cm]
    \tikzstyle{rfc} = [fc, fill=red!5, draw=red, text=red!40!black]       % reset fc
    \tikzstyle{zfc} = [fc, fill=cgreen!5, draw=cgreen, text=cgreen!40!black] % update fc
    \tikzstyle{nfc} = [fc, fill=blue!5, draw=blue, text=blue!40!black]    % new fc

    \tikzstyle{op}   = [rectangle, draw, semithick, text centered, minimum size=5mm, rounded corners]
    \tikzstyle{rop}  = [op, fill=red!5, draw=red, text=red!40!black]       % reset ops
    \tikzstyle{zop}  = [op, fill=cgreen!5, draw=cgreen, text=cgreen!40!black] % update ops
    \tikzstyle{nop}  = [op, fill=blue!5, draw=blue, text=blue!40!black]    % new ops
    \tikzstyle{hop}  = [op, fill=gray!5, draw=gray, text=gray!40!black]    % hidden ops

    \tikzstyle{line}  = [draw, semithick, ->]   % black line
    \tikzstyle{rline} = [line, color=red]    % reset gate line
    \tikzstyle{zline} = [line, color=cgreen] % update gate line
    \tikzstyle{nline} = [line, color=blue]   % new gate line
    \tikzstyle{hline} = [line, color=gray]  % hidden gate line
    \tikzstyle{oline} = [draw, line width=0.8mm, color=white] % line for overlaps
    \tikzstyle{grubox} = [draw, very thick, color=orange, fill=white] % outer box line

    % Labels for inputs and previous state
    \node [inout, align=right] (htm1) {$h_{t-1}$};
    
    % Nodes
    \node [rfc, right=8mm of htm1] (Whr) {$W_{hr}$};
    \node [zfc, below=of Whr]  (Whz) {$W_{hz}$};
    \node [nfc, below=of Whz]  (Whn) {$W_{hn}$};
    
    \node [rfc, below=2mm of Whn]  (Wir) {$W_{ir}$};
    \node [zfc, below=of Wir] (Wiz) {$W_{iz}$};
    \node [nfc, below=of Wiz] (Win) {$W_{in}$};

    \coordinate (xt_coord) at ($(Wir -| htm1)$);
    \node at (xt_coord) [inout, align=right] (xt) {$x_t$};

    \node [rop, right=9mm of Wir] (r_plus) {$+$};
    \node [zop, right=of Wiz] (z_plus) {$+$};
    \node [nop, right=of Win] (n_plus) {$+$};
    
    \node [rop, right=of r_plus] (r_sig) {$\sigma$};
    \node [zop, below=of r_sig] (z_sig) {$\sigma$};
    \node [nop, below=of z_sig] (n_tanh) {$\tanh$};

    \node [nop, right=of r_sig] (r_mul) {$\odot$};
    \node [hop, right=1.3cm of z_sig] (z_compl) {$1-x$};
    \node [hop, right=of z_compl] (n_mul) {$\odot$};
    \coordinate (h_mul_coord) at ($(Whr -| z_compl)$);
    \node at (h_mul_coord) [hop] (h_mul) {$\odot$};
    \coordinate (h_plus_coord) at ($(Whr -| n_mul)$);
    \node at (h_plus_coord) [hop] (h_plus) {$+$};

    \node [inout, right=6mm of h_plus] (ht) {$h_t$};

    % border
    \begin{scope}[on background layer]
        \draw [grubox, rounded corners] ([xshift=5mm,yshift=8mm]htm1.north east) rectangle ([xshift=-5mm,yshift=-4.2cm]ht.south west);
    \end{scope}

    % Edges
    \draw [oline] (htm1.east) -- (Whr.west);
    \draw [line] (htm1.east) -| ++(8mm,5mm) |- ++(0cm,0cm) -| (h_mul.north);
    \draw [line] (htm1.east) -- (Whr.west);
    \draw [line] (htm1.east) -| ++(8mm,0cm) |- (Whz.west);
    \draw [line] (htm1.east) -| ++(8mm,0cm) |- (Whn.west);
    
    \draw [oline] (xt.east) -- (Wir.west);
    \draw [line] (xt.east) -- (Wir.west);
    \draw [line] (xt.east) -| ++(9mm,0mm) |- (Wiz.west);
    \draw [line] (xt.east) -| ++(9mm,0mm) |- (Win.west);
    
    \draw [rline] (r_plus.east) -- (r_sig.west);
    \draw [zline] (z_plus.east) -- (z_sig.west);
    \draw [nline] (n_plus.east) -- (n_tanh.west);
    
    \draw [rline] (Whr.east) -| (r_plus.north);
    \draw [zline] (Whz.east) -| (z_plus.north);
    \draw [oline] (Whn.east) -| (r_mul.north);
    \draw [nline] (Whn.east) -| (r_mul.north);

    \draw [oline] (Wir.east) -- (r_plus.west);
    \draw [rline] (Wir.east) -- (r_plus.west);
    \draw [zline] (Wiz.east) -- (z_plus.west);
    \draw [nline] (Win.east) -- (n_plus.west);
    
    \draw [rline] (r_sig.east) -- (r_mul.west) node [midway] {$r_t$};
    \draw [nline] (r_mul.south) |- ++(0mm,-1.7cm) -| (n_plus.south);
    \draw [oline] (z_sig.east) -- (z_compl.west);
    \draw [zline] (z_sig.east) -- (z_compl.west) node [near start] {$z_t$};
    \draw [zline] (z_sig.east) -| ++ (1.3cm,0cm) |- (h_mul.west);
    \draw [oline] (n_tanh.east) -| (n_mul.south);
    \draw [nline] (n_tanh.east) -| (n_mul.south) node [near start] {$n_t$};
    
    \draw [hline] (z_compl.east) -- (n_mul.west);
    \draw [hline] (h_mul.east) -- (h_plus.west);
    \draw [hline] (n_mul.north) -- (h_plus.south);
    \draw [oline] (h_plus.east) -- (ht.west);
    \draw [line] (h_plus.east) -- (ht.west);

    \draw [line, dotted] (ht.north) |- ++(0cm,1.4cm) -| (htm1.north);
    % \node [fill=white, up] [near end] {at new time step};

\end{tikzpicture}
  \caption{A GRU with its building blocks exposed; the different colors denote the operations associated with \cref{eq:gru}.}
  \label{fig:gru}
  \Description{Diagram showing the architecture of a GRU cell. The cell features two the previous hidden state and the current input as inputs and the current hidden state as output. All inputs go through three separate dense layers; each pair of dense outputs is then summed element-wise and passed through a sigmoid activation. These signals correspond to the reset gate, update gate, and candidate activation. In the latter, the input portion of the signal is modulated by the reset gate and the activation function is tanh. The cell output is the sum of the current hidden state and the candidate activation, both modulated by the update gate and its complement. The GRU cell is also fully described by equations 1.}
\end{figure}

GRUs are a simplified version of LSTMs, which implement only two gates and use a single state vector instead of two.
This is achieved through the following equations (based on the formulation in \cite{chung_empirical_2014}):
\begin{subequations}\label{eq:gru}
  \begin{align}
    r_t &= \sigma(W_{ir} x_t + b_{ir} + W_{hr} h_{(t-1)} + b_{hr}) \label{eq:gru1} \\
    z_t &= \sigma(W_{iz} x_t + b_{iz} + W_{hz} h_{(t-1)} + b_{hz}) \label{eq:gru2} \\
    n_t &= \tanh(W_{in} x_t + b_{in} + r_t \odot (W_{hn} h_{(t-1)}+ b_{hn})) \label{eq:gru3} \\
    h_t &= (1 - z_t) \odot n_t + z_t \odot h_{(t-1)} \label{eq:gru4}
  \end{align}
\end{subequations}
where $x_t$ is the input at the current time step, $h_{(t-1)}$ is the previous hidden state.
The gates are the \textit{reset gate} $r_t$ and the \textit{update gate} $z_t$, implemented in \cref{eq:gru1,eq:gru2}, while the \textit{hidden state} $h_t$, which also serves as output, is implemented in \cref{eq:gru4}.
Finally, GRUs also include a \textit{new state} (or \textit{candidate activation}) $n_t$ which is computed in \cref{eq:gru3}, and is used to compute the new hidden state.
The reset gate controls how much of the previous state is kept in memory, while the update gate controls how much of the new state is added to the memory.
The GRU architecture, along with all its building blocks and operations described above, is depicted in \cref{fig:gru}.

\subsection{Quantization}\label{ssec:quant}
In the field of signal processing, quantization is the process of mapping a continuous set of values to a finite one.
When applied to neural networks, quantization is a compression technique aimed at reducing the memory footprint and computational complexity of the model by limiting the precision of its weights and activations.
Since most deep learning models are heavily over-parameterized, there is often a large margin for reducing their representational precision without significantly affecting performances~\cite{gholami_survey_2021}.

In this work, we consider linear (or uniform) quantization, where the original real values are compressed into finite sets of evenly-spaced quantization levels.
What follows is a brief overview of the technique; for a more comprehensive treatment please refer to \cite{nagel_white_2021,gholami_survey_2021,jacob_quantization_2017}.
Linear quantization is achieved by mapping a real value $r$ to its corresponding quantized $q$ using the following transforms:
\begin{subequations}\label{eq:quant1}
  \begin{align}
    q &= \left\lfloor\frac{r}{S}\right\rfloor - Z \label{eq:quant1a} \\
    r \approx \tilde{r} &= S (q + Z) \label{eq:quant1b}
  \end{align}
\end{subequations}
where $\tilde{r}$ is the dequantized real value, $S$ is the scaling factor, and $Z$ is the zero-point, which are computed as follows:
\begin{equation}\label{eq:quant2}
    S = \frac{\beta - \alpha}{2^b - 1} \quad;\quad
    Z = \left\lfloor\frac{2^b - 1 - \beta}{S}\right\rfloor
\end{equation}
where $b$ is the bit-width of the quantization, and $\alpha$ and $\beta$ are the lower and upper bounds of the clipping range, respectively.
Choosing the clipping range is a crucial step in quantization, as it determines the range of representable values; a straightforward choice is often $\alpha = \min(r)$ and $\beta = \max(r)$.

When $\alpha \neq -\beta$ this is known as \textit{asymmetric quantization}, since the quantization range is not symmetric around zero.
Conversely, when the clipping range is centered around zero, i.e. $\alpha = -\beta$, this is known as \textit{symmetric quantization}.
When using symmetric quantization, we have that $Z = 0$, which simplifies \cref{eq:quant1}.
While asymmetric quantization is more expressive due to the tighter clipping range, symmetric quantization is preferred due to its lower computational overhead.
Oftentimes, in practice, weights are quantized symmetrically while activations are quantized asymmetrically \cite{gholami_survey_2021}.

The quantization parameters $S$ and $Z$ can be computed by simply observing the minimum and maximum values of weights and activations over a representative calibration dataset: this is called \textit{Post-Training Quantization} (PTQ).
Alternatively, they can be computed during \textit{Quantization Aware Training} (QAT) by adding quantization and dequantization nodes to the model graph --- namely implementing \cref{eq:quant1} --- and training the model with the quantization nodes in place.
In this latter case, since the quantization step in \cref{eq:quant1a} is not differentiable, the \textit{Straight-Through Estimator} (STE) \cite{bengio_estimating_2013} is used to backpropagate the gradients through the quantization nodes.
This additional step allows the model to learn to adapt to the quantization, which often results in better performances than PTQ.
In this work, we evaluate our method on both quantization strategies, enabling us to fully appreciate its versatility.

\subsection{Genetic Algorithms}\label{ssec:ga}
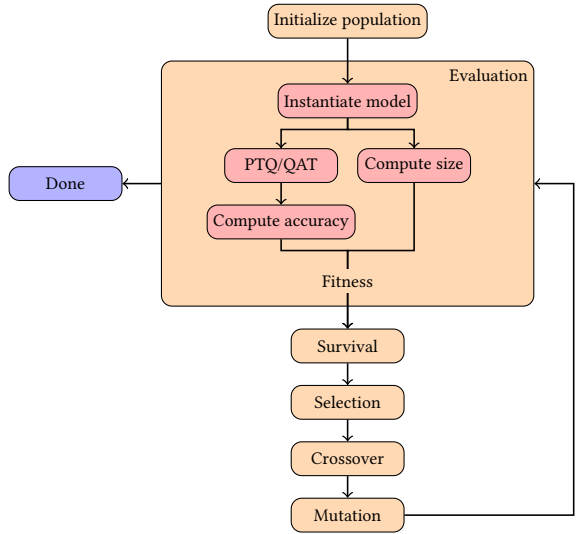
\begin{figure}[ht]
  \centering
  \begin{tikzpicture}[node distance=1cm, auto, scale=0.75, every node/.style={scale=0.75}]

\tikzset{
    pbox/.style={rectangle, rounded corners, minimum width=2cm, minimum height=6mm, text centered, draw=black, fill=orange!30},
    ebox/.style={pbox, fill=red!30},
    arrow/.style={draw, semithick, ->},
}

\node (init) [pbox] {Initialize population};

\node (init_model) [ebox, below=6mm of init] {Instantiate model};
\node (comp_size) [ebox, below right=4mm and -8mm of init_model] {Compute size};
\node (finetune) [ebox, below left=4mm and -8mm of init_model] {PTQ/QAT};
\node (comp_acc) [ebox, below of=finetune] {Compute accuracy};

\node (surv) [pbox, below=2.8cm of init_model] {Survival};
\node (select) [pbox, below of=surv] {Selection};
\node (cross) [pbox, below of=select] {Crossover};
\node (mut) [pbox, below of=cross] {Mutation};

\begin{scope}[on background layer]
    % Define corner points with shifts
    \coordinate (A) at ([xshift=3.3cm,yshift=-4mm]init.south);
    \coordinate (B) at ([xshift=-3.3cm,yshift=4mm]surv.north);

    \path let \p1 = ($(A)-(B)$), \n1 = {veclen(\x1,\y1)} in
          node (eval) [pbox, minimum width=abs(\x1), minimum height=abs(\y1), anchor=center] at ($(A)!0.5!(B)$) {};
\end{scope}

\node (done) [pbox, left of=eval, xshift=-4cm, fill=blue!30] {Done};

\draw [arrow] (init) -- (init_model);
\draw [arrow] (init_model.south) |- ++(0,-2mm) -| (comp_size.north);
\draw [arrow] (init_model.south) |- ++(0,-2mm) -| (finetune.north);
\draw [arrow] (finetune) -- (comp_acc);
\draw [arrow] (comp_size.south) |- ++(0,-1.2cm) -| (surv.north);
\draw [arrow] (comp_acc.south) |- ++(0,-2mm) -| (surv.north);

\draw [arrow] (surv) -- (select);
\draw [arrow] (select) -- (cross);
\draw [arrow] (cross) -- (mut);
\draw [arrow] (mut.east) -- ++(3cm,0) |- (eval.east);
\draw [arrow] (eval.west) -- (done.east);

\node [align=left, anchor=base west, below=2cm of init_model, fill=orange!30] {Fitness};
\node [anchor=north east, minimum height=5mm] at (eval.north east) {Evaluation};

\end{tikzpicture}
  \caption{Typical flow of a genetic algorithm, with red blocks indicating the custom evaluation pipeline.}
  \label{fig:ga}
  \Description{A flowchart representing a genetic algorithm. The first step is to initialize a population of potential solutions with random genes. Then, the fitness of each individual is evaluated. This evaluation step involves several sub-steps, namely the initialization of a model, quantization using PTQ/QAT, and evaluation based on accuracy and model size. These two metrics comprise the aforementioned fitness. The following steps represent the core of the genetic algorithm algorithm. During survival, the population is sorted based on fitness, and the top individuals are kept. During selection, individuals are picked for mating. In Crossover, their genome is combined according to a specific crossover operator. Finally, in mutation, the genome undergoes random variations. The algorithm then loops back to the fitness evaluation. This process is repeated until a stopping criterion is met}
\end{figure}

Genetic algorithms \cite{goldberg2013genetic} are a class of optimization metaheuristics based on the principles of natural selection and biological evolution.
They mimic the logic of biological and genetic adaptation of individuals (solutions) to an environment (search space) and are extensively used in a large variety of domains to optimize multi-dimensional functions that are not differentiable, where classical gradient descent algorithms are not adequate. 
Starting from an initial population, the algorithm iterates over the following phases, illustrated in \cref{fig:ga}: evaluation, survival, selection, crossover and mutation, until an exit condition is met.
A fundamental aspect of GA is the \textit{genome}, a set of parameters representing a potential solution to the given optimization problem, as well as the basis for the crossover and mutation phases.
In this context, each parameter is a \textit{gene}.
Genomes can be encoded in a variety of ways, including binary strings, real-valued vectors, or more sophisticated data structures allowing for discrete or continuous genes.
Due to their reliance on a population of solutions, GA may be computationally demanding but are also suitable for parallel computing.

There exist many variants of GA. 
The Non-dominated Sorting Genetic Algorithm (NSGA) uses a vector fitness function, making it suitable for multi-objective optimization problems. 
In the specific case of NSGA-II \cite{deb2002a}, which we use in our work, a two-dimensional vector and selection heuristic based on the concept of Pareto efficiency (or optimality) is used. 
According to it, a solution whereby all the objectives are worse than those of another solution is said to be ``dominated''.
Conversely, a solution that fares better than the alternatives by at least one objective is considered to be ``non-dominated''.
In the selection phase, the solutions are sorted based on non-dominated sorting and crowding distance. 
This approach has already been used in the context of NAS~\cite{lu_nsganet_2019} and presents itself as an adequate solution when there is a need to optimize a neural network on the basis of different objectives, such as accuracy but also other performance or efficiency-related objectives like latency, model size, and memory footprint.

\section{Modular quantization}\label{sec:modular_quant}
In this section, we present our approach to integer quantization of GRU layers, as well as a method for exploring the search space of possible bit-widths using the GA described in \cref{ssec:ga}.

\subsection{Quantization scheme for GRU}\label{ssec:gru_quant}
Building upon the methods described in \cref{ssec:quant}, we propose a linear quantization scheme for GRUs where each operation is quantized independently, allowing for heterogeneous bit-widths.

By inspecting \cref{fig:gru}, we can see that a GRU layer is composed of 4 main types of operations: linear/dense layers, element-wise additions, element-wise multiplications, and non-linear activation functions.
For each operation, we can define a quantization scheme by applying \cref{eq:quant1b} to the real-valued parameters involved in the operation definition --- namely its inputs, outputs, and relevant weights --- and then solving for the quantized output.

To accommodate integer-only hardware, we further restrict the quantization to integer-only computation.
This is done by representing the combined scaling factors $M_{\ast}$ in \cref{eq:lin_quant,eq:sum_quant,eq:prod_quant} as fixed-point with an $n$-bit fractional part, so that $M_{\ast} \approx 2^{-n} M_0$ (where ${}_{\ast}$ indicates any of the specific qualifiers introduced in the aforementioned equations).
This scaling can then be trivially implemented with a multiplication and a shift operation.

In the following subsections, we elaborate on how each of the operations involved in a GRU layer is quantized.

\subsubsection{Linear layer}
A linear (or dense) layer is defined as:
\begin{equation}\label{eq:lin}
  \boldsymbol{y} = \boldsymbol{W}\boldsymbol{x} + \boldsymbol{b}
\end{equation}
where $\boldsymbol{x}$ is the input vector, $\boldsymbol{W}$ is the weight matrix, $\boldsymbol{b}$ is the bias vector, and $\boldsymbol{y}$ is the output vector.
By applying \cref{eq:quant1b} to each of the parameters, we obtain:
\begin{equation}
  S_y(q_y - Z_y) = S_w(q_w - Z_w) S_x(q_x - Z_x) + S_b(q_b - Z_b)
\end{equation}

Subsequently, we can assume symmetric quantization for weights and biases ($Z_w = Z_b = 0$), and adopt a shared scaling factor for matrix multiplication and addition ($S_b = S_w S_x$). Applying this and solving for the quantized output $q_y$ yields:
\begin{align}
  S_y(q_y - Z_y) &= S_w S_x(q_w q_x - Z_x q_w + q_b) \\
  q_y &= \frac{S_w S_x}{S_y} (q_w q_x - Z_x q_w + q_b) + Z_y
  % q_y &= M (q_w q_x - Z_x q_w + q_b) + Z_y \\
\end{align}

We further simplify computation by substituting the following:
\begin{equation}
  \frac{S_w S_x}{S_y} = M \quad;\quad
  q_b - Z_x q_w = q_{\text{bias}}
\end{equation}
where $M$ is the combined scaling factor (representable as a fixed-point number as mentioned earlier) and $q_{\text{bias}}$ is a pre-computed term.
This finally gives us:
\begin{equation}\label{eq:lin_quant}
  q_y = M (q_w q_x + q_{\text{bias}}) + Z_y
\end{equation}

\subsubsection{Element-wise sum}
When quantizing an element-wise sum, we similarly apply \cref{eq:quant1b} to each of the parameters involved in the sum, and then solve for the quantized output:
\begin{align}
  \boldsymbol{y} &= \boldsymbol{x_1} + \boldsymbol{x_2} \\
  S_y(q_y - Z_y) &= S_1(q_1 - Z_1) + S_2(q_2 - Z_2) \\
  q_y &= \frac{S_1}{S_y} (q_1 - Z_1 + \frac{S_2}{S_1} (q_2 - Z_2)) + Z_y
\end{align}

We are now left with two combined scaling factors, $M_{\alpha} = \sfrac{S_1}{S_y}$ and $M_{\beta} = \sfrac{S_2}{S_1}$. 
Notice how $M_{\beta}$ contributes to matching the quantization grid of the two operands, while $M_{\alpha}$ matches the final output range.
After substituting the scaling factors, we obtain:
\begin{equation}\label{eq:sum_quant}
  q_y = M_{\alpha} (q_1 - Z_1 + M_{\beta} (q_2 - Z_2)) + Z_y
\end{equation}

\subsubsection{Element-wise product}
We once again apply the same familiar process as above:
\begin{align}
  \boldsymbol{y} &= \boldsymbol{x_1} \odot \boldsymbol{x_2} \\
  S_y(q_y - Z_y) &= S_1(q_1 - Z_1) \odot S_2(q_2 - Z_2) \\
  q_y &= \frac{S_1 S_2}{S_y}(q_1 q_2 - q_1 Z_2 - q_2 Z_1 + Z_1 Z_2) + Z_y
\end{align}

This time, after some simplification, we are left with a single scaling factor $M_{\gamma} = \sfrac{S_1 S_2}{S_y}$:
\begin{equation}\label{eq:prod_quant}
  q_y = M_{\gamma} (q_1 q_2 - q_1 Z_2 - q_2 Z_1 + Z_1 Z_2) + Z_y
\end{equation}

\subsubsection{Activations}
Finally, we consider the quantization of the sigmoid and hyperbolic tangent activation functions, here both denoted as $\sigma$.
In this case, we build lookup tables (LUT), according to the following process:
\begin{enumerate}
  \item Generate a vector of length $L$ of quantized input values $q_x$ between $0$ and $2^b - 1$, where $b$ is the given bit-width;
  \item Compute its real-valued equivalent $\tilde{x}$ using \cref{eq:quant1b};
  \item Compute the real-valued activation $\tilde{y} = \sigma(\tilde{x})$;
  \item Quantize $\tilde{y}$ using \cref{eq:quant1a} to obtain the LUT $q_y$.
\end{enumerate}

During inference, $\left\lfloor \sfrac{L}{2^b - 1} \right\rfloor \cdot q_x$ is used as the table index, and the corresponding quantized output $q_y$ is returned.

\subsection{Quantization scheme search}\label{ssec:search}
As mentioned in \cref{ssec:ga}, genetic algorithms operate on a genome representing each potential solution. 
In the context of our optimization problem, the genome is a vector of $17$ genes, each representing the bit-width of a quantized operation in a GRU cell (i.e., each of the blocks in \cref{fig:gru}), encoded as an integer value in the range between $2$ and $8$.
The combinations of genes and their bit-width values correspond to the search space of our problem.

We formulate our search task as a multi-objective maximization problem, where the objectives under consideration are the model accuracy and $\widehat{\mathbb{M}}^{\mathsf{c}}$, i.e. the normalized complement of the model size, defined as follows:
\begin{equation}
  \widehat{\mathbb{M}}^{\mathsf{c}} = 1 - \frac{\mathbb{M}_{\text{Q}}}{\mathbb{M}_{\text{FP16}}}
\end{equation}
where $\mathbb{M}_{\text{Q}}$ is the quantized model size and $\mathbb{M}_{\text{FP16}}$ is the model size of the floating-point baseline, both in bits.
The model sizes $\mathbb{M}_{\ast}$ are computed as:
\begin{equation}
  \mathbb{M}_{\ast} = \sum_{\phi \in \Phi} \text{numel}(W^{\phi}) \cdot {N_b}^{\phi} + \text{numel}(b^{\phi}) \cdot 32
\end{equation}
where $\Phi$ is the set of all linear layers in the model and $W^{\phi}$, $b^{\phi}$, and ${N_b}^{\phi}$ are the weight matrix, bias vector, and chosen bit-width for the given layer $\phi$, respectively; finally, $\text{numel}()$ is the number of elements in its argument.
Note that the bit-width of the bias vector is fixed to $32$ bits, which we assume to be the size of the accumulator register.

The overall genetic search workflow, along with the aforementioned evaluation pipeline, is depicted in \cref{fig:ga}.
During the survival stage, a rank and crowding strategy is adopted, whereby solutions are first sorted into fronts based on non-domination, assigning lower ranks to Pareto-optimal solutions. 
To maintain diversity, preference is given to solutions with higher crowding distance, which measures the proximity of solutions within the same front; this ensures an even distribution across the Pareto front. 
Subsequently, mating candidates are selected through binary tournaments, based on their ranks and crowding distances, and new solutions are generated by means of Simulated Binary Crossover and Polynomial Mutation~\cite{deb2007self}.
These two are crossover and mutation strategies, respectively, that are suitable for real-valued genes.

\section{Experimental evaluation}\label{sec:exp_eval}
We conducted a number of experiments to evaluate our methods, spanning 4 different sequential classification tasks: row-wise sequential MNIST, pixel-wise sequential MNIST, keyword spotting with 4 keywords, and keyword spotting with 10 keywords.
In the following subsections, we describe the experimental setup used to conduct the genetic search, along with the details of each task, and discuss the results.

\subsection{Setup}\label{ssec:exp_setup}
Each task was performed using a simple recurrent model with a single GRU layer followed by a linear layer.
For each task, we trained an FP16 model and quantized it homogeneously, using either PTQ or QAT, with bit-widths ranging from 3 to 8; 
we use these models as our baselines.
During training, we minimize the cross-entropy loss between the model predictions and the ground truth labels.
Subsequently, we used NSGA-II to search for optimal mixed-precision quantization schemes.
\cref{tab:train_hyperparams} provides an overview of the hyperparameters used during FP16 training, homogeneous quantization, and mixed-precision quantization (through genetic search).

\begin{table}[t]
  \caption{Hyper-parameters for model training; mixed-precision values correspond to those used for QAT.}
  \label{tab:train_hyperparams}
    \begin{tabular}{l|ccc}
      \toprule
      Parameter     & FP16      & Homogeneous        & Mixed-precision \\
      \midrule
      Batch size    & $256$     & $1024/2048$        & $1024$ \\
      Epochs        & $50$      & $30$               & $12$   \\
      Train. split  & $100\%$   & $100\%$            & $10\%$ \\
      Valid. split  & $5\%$     & $5\%$              & $5\%$  \\
      Valid. every  & $5$       & $5$                & $3$    \\
      Optimizer     & Adam~\cite{Kingma2015}      & Adam               & Adam    \\
      Learning rate & $10^{-3}$ & $5 \times 10^{-5}$ & $5 \times 10^{-5}$ \\
      \bottomrule
    \end{tabular}
\end{table}

During the genetic search, we randomly generated the genome for the initial population by sampling the search space --- i.e., by generating a $17$-element vector of uniformly-distributed integers in the closed interval $[2, 8]$.
The search was terminated after $20$ generations.
This process was repeated for each task.
We use the NSGA-II implementation provided by \textit{pymoo}~\footnote{\url{https://pymoo.org/algorithms/moo/nsga2.html}}.

We then evaluated the models on the test set using the same metrics described in \cref{ssec:search}.
To avoid overfitting on our test set, during the survival and selection phases of the genetic search (see \cref{ssec:ga}), the accuracy of each individual is computed on a separate validation split corresponding to an unseen $10\%$ of data.

\subsection{Sequential tasks}\label{ssec:tasks}
The following subsections describe the datasets, hyperparameters, and training procedures used for each task.

\subsubsection{Sequential MNIST}
For these experiments, we use the MNIST dataset \cite{deng_mnist_2012}, adapted to work for sequential tasks. 
The dataset comprises $28 \times 28$ gray-scale images of handwritten digits, divided into $60\,000$ training and $10\,000$ test samples.

For the row-wise variant, we feed each data point as a sequence of $28$ time steps, each consisting of a $28$-pixel feature vector; thus, each row of pixels is treated as a time step.
For the pixel-wise variant, we feed each data point as a sequence of individual pixels, each treated as a one-dimensional feature vector.
Since training with sequences of length $784$ is exceedingly expensive, we resize the images to $19 \times 19$ pixels using bilinear interpolation before converting them to sequences, resulting in sequences of length $361$.

For this set of experiments, we perform QAT on the models; for an overview of the training hyperparameters, please refer to \cref{tab:train_hyperparams}.
The models feature $128$ and $256$ hidden features for the row-wise and pixel-wise task variants, respectively.

\subsubsection{Keyword spotting}
For these experiments, we use the Speech Commands dataset \cite{warden_speech_2018}, which comprises $65\,000$ one-second audio clips of $30$ different words, divided into $20\,000$ training, $5\,000$ validation, and $10\,000$ test samples.
The input data consists of $80$ Mel-frequency log-power features extracted from the STFT of the audio signal, computed on 32-millisecond windows with $50\%$ overlap, resulting in sequences of $63$ time steps.

Although the original dataset contains $30$ words, for our experiments we only consider the first $4$ and $10$ words, respectively.
This is because, given its relatively simple architecture, the accuracy of the full-precision model on the full dataset was already quite low.

To further test the viability of our quantization scheme and parameter search, we perform PTQ on these models, instead of QAT.
In this case, we use the entire training set for calibration.
Similarly to the previous experiments, the models feature $128$ and $256$ hidden features for the $4$-word and $10$-word task variants.

\subsection{Results}
\begin{figure*}[t]
  \centering
  \includegraphics[width=\linewidth]{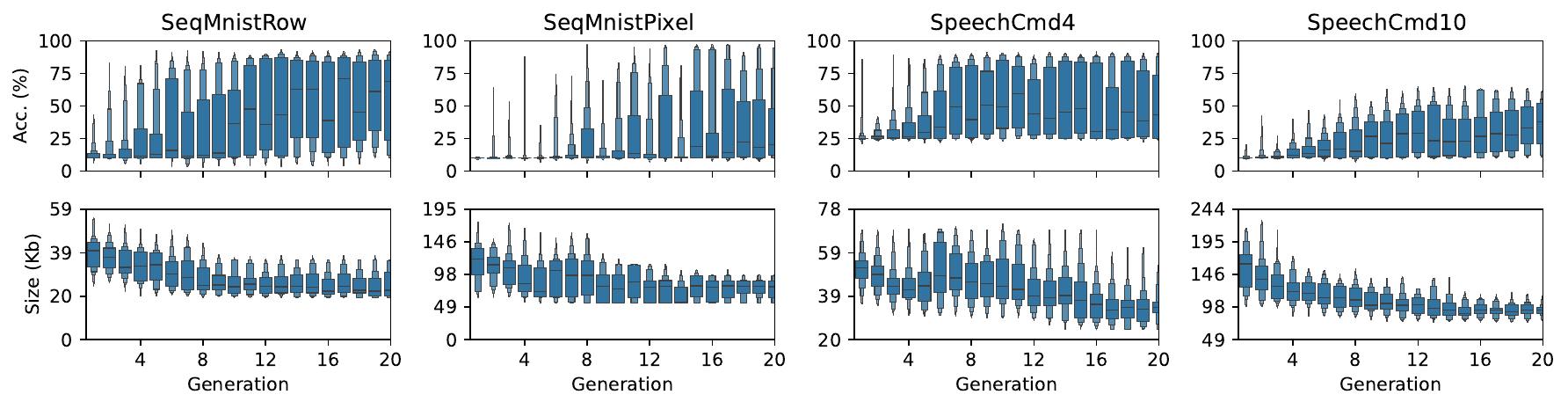}
  \caption{Enhanced box plot showing the distributions of accuracy and model size across generations for the given tasks.}
  \label{fig:res_gen}
  \Description{Enhanced box plot showing the distributions of accuracy and model size. The figure features 8 subplots, divided into two rows and four columns. Each column is a classification task, the first row shows accuracy, and the second shows model size in kilobytes; on each subplot, the x-axis corresponds to the generation, with the leftmost box meaning oldest and the rightmost box meaning latest. Each subplot shows median, inter-quantile ranges as increasingly thinner boxes, and outliers. The plots show a generalized upward trend in accuracy and a downward trend in model size, over generations.}
\end{figure*}

\cref{fig:res_gen} shows their progressive improvement over each generation.
As we can see, the model accuracy generally increases over time, while the model size decreases.
This trend can be observed in all four experiments, although the rate of improvement decreases steeply over time, and the maximum accuracy is often reached early on in the process.
Indeed, in all the experiments besides \textit{SpeechCmd10}, the median accuracy across generations oscillates dramatically while maintaining an upward trend, and we observe a relatively large variance for accuracy figures.
In \textit{SpeechCmd10}, the accuracy continues to increase up to the very last generations, hinting that extending the genetic search could have further improved our results.
Conversely, the model size progressively decreases and converges in all experiments except \textit{SpeechCmd4}, where a larger variance is observed.
These observations are likely explained by the lack of inequality constraints on the objectives, causing lightweight but low-accuracy solutions to survive.
Similarly, the absence of an \textit{exploitation} phase prevents our search to further refine our high-quality specimens.

\begin{figure}[t]
  \centering
  \includegraphics[width=\linewidth]{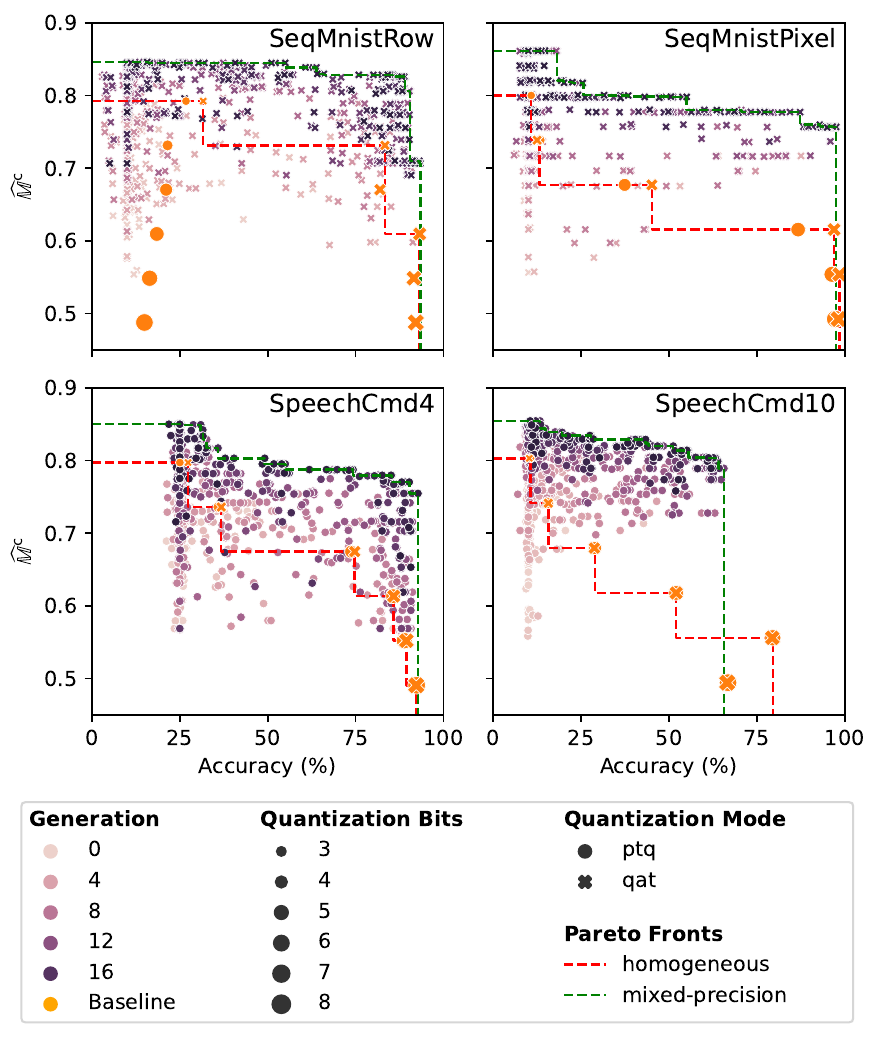}
  \caption{Pareto fronts (higher is better) for homogeneous (baseline) vs. mixed-precision (genetic search) quantization; each dot is a solution/individual.}
  \label{fig:res_pareto}
  \Description{Scatter-plots of model performances, with one subplot for each classification task. Each dot is a fine-tuned model stemming from either a baseline or a solution. The x-axis represents accuracy; the y-axis represents normalized model size complement as described in equation 8. For homogeneous quantization, the dot size represents the number of quantization bits; for mixed-precision quantization, the dot shade represents its generation; baseline models are represented by orange dots. On each subplot, the Pareto front derived by the homogeneous quantization is shown in red and the one based on mixed-precision is shown in green; in all but one subplots, the green Pareto front dominates the red one.}
\end{figure}

\cref{fig:res_pareto} shows the Pareto fronts for the search, for each of the four experiments.
When comparing the Pareto front projected by the homogeneously-quantized baseline models with the one obtained by the mixed-precision search, we can see that the latter consistently dominates the former, with the exception of the 8-bit QAT/PTQ models, which achieve better accuracy than any of the heterogeneously-quantized models.
In all the other cases, however, we achieve better or comparable accuracy and lower model size with our method.
We can also see that there appear lots of mixed-precision solutions with chance-like accuracy, presenting as vertical stripes on the left side of the plot.
However, these often comprise points from the earliest generations, further proving the effectiveness of the genetic search. 
While portions of our Pareto front present solutions with very low model size, oftentimes these provide very low accuracy, making them potentially unsuitable to fulfill the given task; nevertheless, they could be employed as part of an ensemble model, leveraging parallelization.

\begin{figure}[ht]
  \centering
  \includegraphics[width=\linewidth]{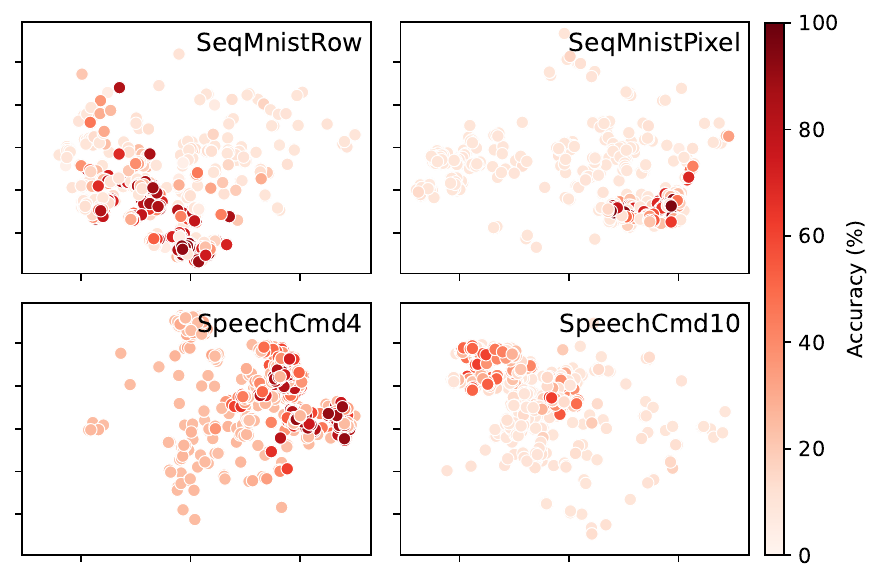}
  \caption{Dimensionally-reduced visualization (using t-SNE) of solutions' genomes, colored by accuracy.}
  \label{fig:res_clusters}
  \Description{Scatter-plot of dimensionally-reduced quantization scheme, with one subplot for each classification task. Each axis represents a non-linear combination of quantization bits for each building block of the GRU cell, obtained using the t-SNE algorithm. The dots represent individual mixed-precision solutions derived using the genetic search, and their color corresponds to the evaluated accuracy. On each subplot, the solutions with the highest accuracy are confined inside a limited region, hinting at the fact that these high-performing models share similar properties.}
\end{figure}

Furthermore, we tried to verify whether the discovered optimal solutions shared any characteristics that could be used to inform the network design and quantization.
To do this, we fit a t-SNE dimensionality reduction model on the bit-widths associated with each building block of the GRU and subsequently color the dots according to the accuracy achieved.
The emergence of clusters, as shown in \cref{fig:res_clusters}, is evidence that there are some shared quantization patterns that are common among the best-performing models. 
Most notably, these clusters of high-performing solutions seem to occupy different regions of the t-SNE space, validating the need for bespoke quantization.
Intuitively, this could be motivated by the different requirements of each task, e.g. problems featuring long sequences require more emphasis, and therefore higher resolution, on the model's internal state.

\section{Conclusion}
In this work, we presented a novel quantization scheme for GRUs and used Genetic Algorithms to simultaneously optimize for accuracy and model size by selecting the appropriate bit-width of each operation.
Based on our preliminary results on a variety of simple sequence classification tasks, the mixed-precision solutions achieve better Pareto efficiency for our chosen metrics.
Namely, in all experiments except one, we achieve a model size reduction between $25\%$ and $55\%$ while maintaining the same (or better) accuracy as the 8-bit homogeneously-quantized model.
Similarly, when considering the relatively small 4-bit homogeneous baselines, we appreciate an increase in accuracy corresponding to between $2$ to $4$ times a similarly-sized heterogeneously-quantized model.

While promising, our solution presents some limitations.
Most notably, to reap the efficiency and inference speed benefits of the heterogeneous quantization scheme, hardware support is essential.
Furthermore, as an extension of this work, the genetic search could be enhanced by introducing an additional \textit{exploitation} phase and by constraining our objectives to useful regions of the Pareto front, as mentioned in \cref{sec:exp_eval}.

It remains to be determined whether our solution would scale to more challenging tasks such as speech enhancement, which would require a more complex model architecture and a larger dataset; these drawbacks could be ameliorated with knowledge distillation and dataset distillation techniques, respectively.
Finally, it is worth noting that the quantized operators described here, along with the genetic search for optimal bit-widths, could also be applied to LSTMs; we leave this investigation for future work.

\begin{acks}
We thank E. Njor for sharing his knowledge on NAS and R. James for the help in brainstorming the initial idea.
This work has received funding from the European Union’s Horizon research and innovation programme under grant agreement No 101070374.
\end{acks}

\balance
\bibliographystyle{ACM-Reference-Format}
\bibliography{refs}

% \appendix
\end{document}